\documentclass[letterpaper, 10 pt, conference]{ieeeconf}  % Comment this line out if you need a4paper

\IEEEoverridecommandlockouts                              % This command is only needed if 
\usepackage{graphics} % for pdf, bitmapped graphics files
\usepackage{epsfig} % for postscript graphics files
\usepackage{amsmath} % assumes amsmath package installed
\usepackage{amssymb}  % assumes amsmath package installed
\usepackage[ruled]{algorithm2e}
\usepackage{tabularx}
\usepackage[table]{xcolor}
\usepackage{hhline}
\usepackage{multirow}
\usepackage{listings}
\usepackage{color}
\usepackage{url}
\usepackage{hyperref}
\usepackage{authblk}

\title{\LARGE \bf
Automatic Design of Task-specific Robotic Arms
}

\author[1]{Ruta Desai\thanks{Author emails: \tt\footnotesize\{rutad, scoros\}@cmu.edu;}}
\author[2]{Margarita Safonova}
\author[1,2]{Katharina Muelling}
\author[1,3]{Stelian Coros}
\affil[1]{Robotics Institute, Carnegie Mellon University}
\affil[2]{National Robotics Engineering Center (NREC)}
\affil[3]{ETH Zurich \thanks{\tt\footnotesize\{rsafonov, kmuelling\}@nrec.ri.cmu.edu}}

\begin{document}

\maketitle
\thispagestyle{empty}
\pagestyle{empty}

%%%%%%%%%%%%%%%%%%%%%%%%%%%%%%%%%%%%%%%%%%%%%%%%%%%%%%%%%%%%%%%%%%%%%%%%%%%%%%%%
\begin{abstract}
We present an interactive, computational design system for creating custom robotic arms given high-level task descriptions and environmental constraints. Various task requirements can be encoded as desired motion trajectories for the robot arm's end-effector. Given such end-effector trajectories, our system enables on-demand design of custom robot arms using a library of modular and reconfigurable parts such as actuators and connecting links. By searching through the combinatorial set of possible arrangements of these parts, our method generates a functional, as-simple-as-possible robot arm that is capable of tracking the desired trajectories. We demonstrate our system's capabilities by creating robot arm designs in simulation, for various trajectory following scenarios.

% In order to automatically create valid robot arm designs using these parts, our system efficiently searches 
% Automatic creation of valid designs is achieved by efficiently searching the 
%space of robot designs that consist of combinatorial set of possible arrangements of these modular parts.

\end{abstract}

%%%%%%%%%%%%%%%%%%%%%%%%%%%%%%%%%%%%%%%%%%%%%%%%%%%%%%%%%%%%%%%%%%%%%%%%%%%%%%%%
\section{Introduction}
Many complex manufacturing scenarios require manipulation in confined spaces, while avoiding obstacles. In particular, specialized domains such as airplane manufacturing involve many such intricate motions, and therefore demand substantial manual labor. Commercially available 6 degrees of freedom (DOF) robotic arm systems such as KUKA~\cite{kuka}, Universal robots~\cite{ur} etc. cater to a large variety of manipulation tasks. However, these robotic arms cannot be modified for specialized tasks, nor can they be reconfigured when the task requirements change. Instead, robotic arms designed using reconfigurable, modular parts can be adapted as per one's need. Therefore, recent commercial robotic systems such as HEBI robotics~\cite{hebi}, Modbot~\cite{modbot} have started promoting the use of customizable designs by providing modular building parts. Given such a modular part library, our goal is to enable designers and engineers to quickly create valid robotic arm designs for the task at hand. Towards this goal, we present a computational approach that automatically generates valid designs given user-specified task requirements.

We formulate the automatic robot design as a search problem through the space of all possible arrangements of the modular parts. To find valid design solutions efficiently, we encode the task requirements with a heuristics, and leverage it to guide the search towards appropriate designs. Our automatic synthesis approach is further supported by a visual and interactive design interface that allows users to intuitively specify their task requirements in terms of desired robot arm motion. Users can further test the auto-generated designs in a physics-based simulation within our system. We give an overview of our interactive system, and the automatic design generation in Sec.~\ref{sec:system}, and~\ref{sec:search} respectively. We then show various robot arm designs automatically created with our system (Sec.~\ref{sec:results}). Although our current focus is on synthesizing robotic arms, our automatic design approach is generic, and can be applied to a variety of robotic systems.

\begin{figure}[htbp]
	\centering
	\includegraphics[width=\columnwidth]{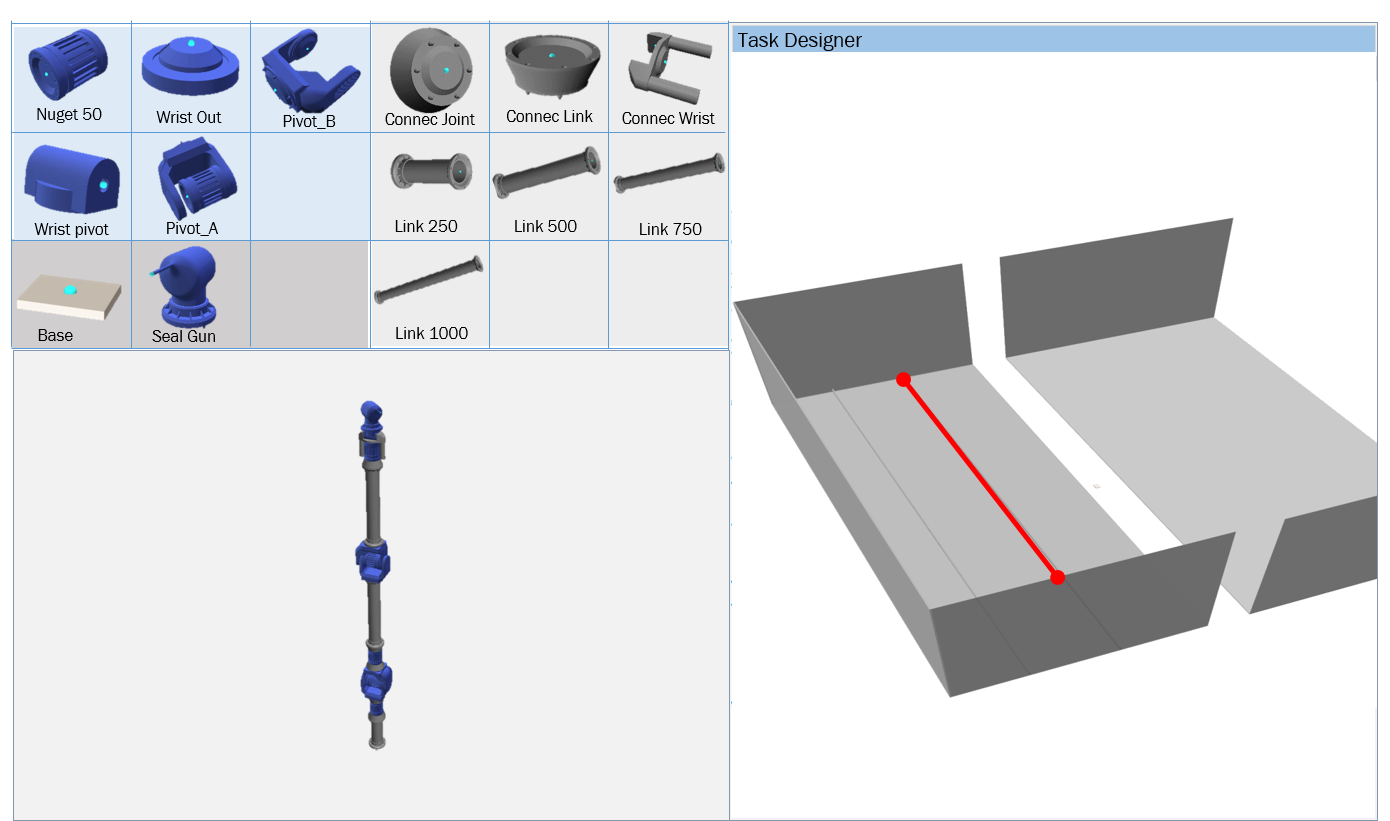}
	\caption{Our graphical interface is shown here. Users can manually or automatically design robots in the design window on the left, using the modular parts displayed in a menu on the top. Actuators are shown in blue, while the connecting parts are highlighted in gray. We used the NREC Modular Actuator (Nugget) for our designs. The task designer on the right is powered by a physics simulation, and allows users to define real-world task scenarios using obstacles (gray) and target trajectories (red). }
	\label{gui}
\end{figure}

% may cater to wide variety of tasks, they cannot be reconfigured when the task requirements change 
%Commercial 6 DOF arms with fixed structure may fall short in their abilities to adapt to 
%The complexity of such motions in specialized domains such as airplane manufacturing therefore demands substantial manual labor.
% one must figure how to configure various parts into a functional design that achieves the desired motions for a task. 
%allows using a modular part library of one's choice for creating custom robot arms.
\section{Related work}
Task-specific manipulator design based on high-level descriptions has received considerable attention in robotics community. For instance, methods to design manipulators that reach desired configurations in a specified workspace while avoiding joint singularities have been proposed~\cite{paredis1991approach,kim2003new,ceccarelli2004multi}. However, most prior methods focus on optimizing continuous parameters, such as limb lengths, for a given robot arm with fixed number of joints. In contrast, our method creates robotic arms entirely from scratch using discrete modular components. To create custom designs with such discrete parts, we build upon our previous work that defines an abstraction for modeling modular components and their connections in a robotic device~\cite{desai2017computational}. Using this abstraction, we also previously developed a search-based design approach that auto-completed structures of user created partial robot designs~\cite{desai2017computational}. In this work, we extend this  framework to automatically create complete designs while accounting for high-level task requirements such as a desired motion for the robot arm's end-effector. 

\section{Graphical design system}
\label{sec:system}
To support user design of custom task-specific robots with a library of modular parts, we develop an intuitive graphical user interface. Along with providing a visual way to interact with our automatic design synthesis algorithm, our interface also enables users to define their task requirements easily. To enable task and robot design simultaneously, our interface is divided into two windows --  a robot design window on the left, and a physics simulation powered task designer on the right (see fig.~\ref{gui}). 

The robot design window displays the components from the part library at the top, and supports manual and automatic design of robotic arms using these parts. The parts include a robot base, actuators, connecting links, and end-effectors. The base part denotes the robot's supporting base and can be used to define the robot arm's position in the world with respect to the environment.
Various end-effectors such as sealing gun, welding machine etc. are provided to create robotic arms that cater to a variety of desired tasks. The task designer allows users to specify target paths for the robot arm to follow, as well as to create an environment with obstacle corresponding to a real-world scenario. Finally, users can test their designs using the physics-based simulation in the task designer before assembling a real design.

 Manual editing is enabled using intuitive drag-and-drop of parts from the menu, and is adapted from our previous work~\cite{desai2017computational}. For automatic design, users provide a robot base position in the real world, and a desired trajectory for the arm to follow. Users can also specify a desired end-effector such as sealing gun or a welding machine, according to their task requirements. Based on these specifications, our algorithm automatically generates a valid arm design. 

\section{Automatic design using informed tree-search}
\begin{figure}[htbp]
	\centering
	\includegraphics[width=\columnwidth]{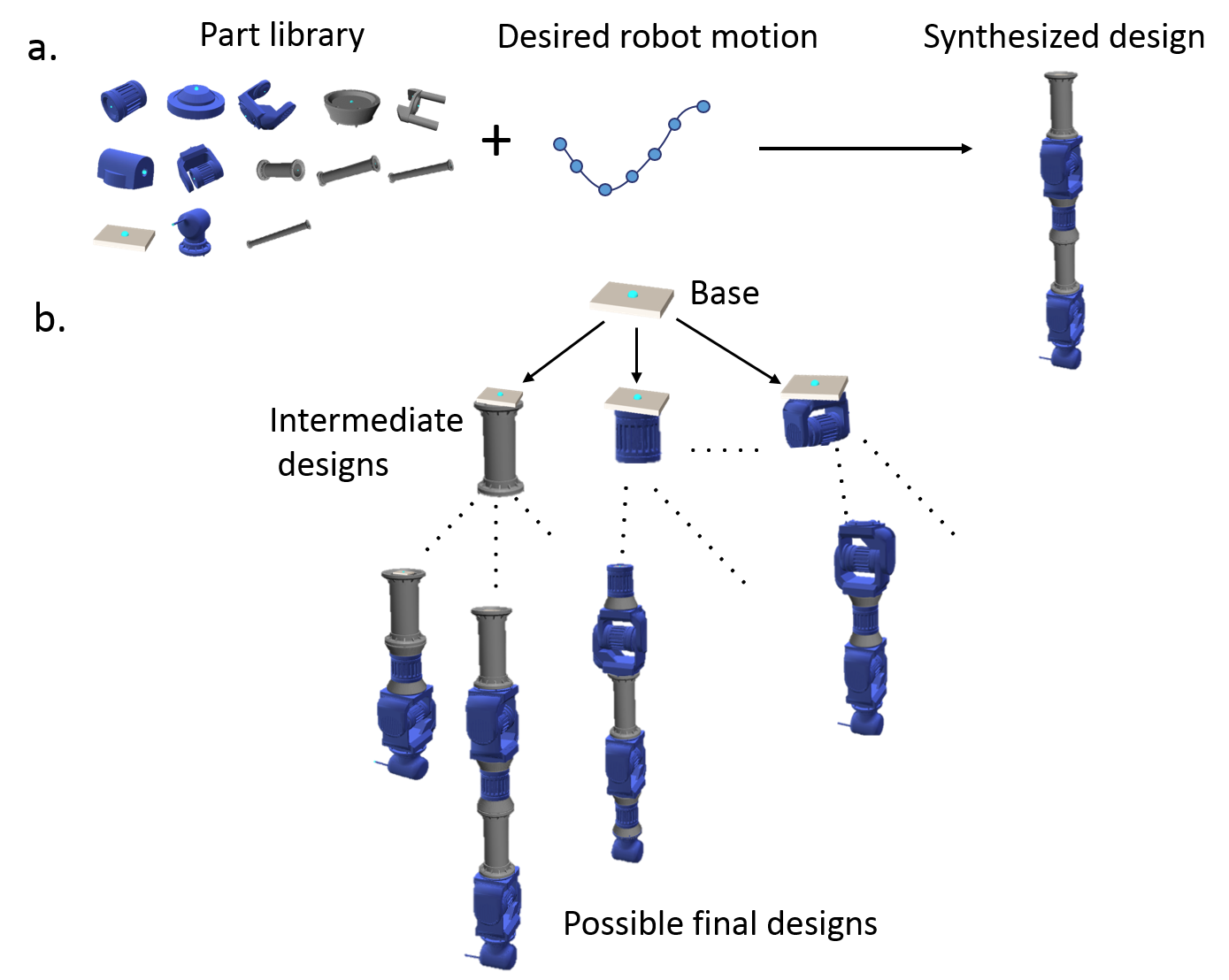}
	\caption{(a)~Given a library of modular parts and a desired robot motion, our system automatically synthesizes valid robotic arm designs.~(b)~Automatic design is formulated as a search problem over recursively created tree of all possible designs.}
	\label{overview}
\end{figure}

Given a library of modular building parts and a user-specified robot arm end-effector motion corresponding to a task, our design synthesis method aims to generate the simplest robot that can execute this motion (see fig.~\ref{overview}(a)). Our design abstraction, which models modular parts and their connections, allows us to map the input part library to a space of possible robot designs. Specifically, the library parts $p$ are modeled as rigid bodies, and their compatibilities are defined using connection rules $c$\footnote{A connection rule $c$ for two compatible parts $p_1$ and $p_2$ is defined as $c=\{p_1,p_2, {}_1\textbf{T}_{2}\}$, where ${}_1\textbf{T}_{2}$ represents a rigid transformation of $p_2$ relative to $p_1$ at the time of connection.}~\cite{desai2017computational}. A robot design $\mathcal{D}=\{\mathcal{P},\mathcal{C}\}$ can now be represented by a collection of interconnected parts $\mathcal{P}$ and their connections $\mathcal{C}$. Combinatorially many such robot designs consisting of different collections of parts $\mathcal{P}$ can be constructed even with a small-sized part library. To find the simplest valid robot design out of all such possible robot designs, we start with a robot base, and recursively construct new designs with increasing number of components till a valid design is found (see fig.~\ref{overview}(b)).

\label{sec:search}

\subsection{Tree of designs}
The recursive approach for synthesizing new designs is motivated by the following observation. Consider a robot design $\mathcal{D}$ composed of part collections $\mathcal{P}$. One can create new \textit{children} designs of $\mathcal{D}$, each with potentially different motor capabilities, by appending compatible parts from the library to $\mathcal{D}$, as defined by the connection rules. As a result, even if a design $\mathcal{D}$ is not well-suited for a user-specified motion, one of its children designs might be. This synthesis approach can be well-represented with an acyclic graph (tree), where each node in the graph corresponds to a design $\mathcal{D}$. The edge between a parent node $\mathcal{D}_1$ and a child node $\mathcal{D}_2$ describe the addition of a single part to $\mathcal{D}_1$, using corresponding connection rules, for creating $\mathcal{D}_2$. The root of the tree corresponds to a base that supports the robotic arm, and the tree terminates at \textit{goal nodes} that correspond to valid designs, capable of executing user-specified motions (see fig.~\ref{overview}(b)). Since the design space is combinatorial, and the depth of the tree is potentially unbounded (one can keep adding more components), brute force construction and traversal of such a tree of designs is very expensive. Instead, we leverage an existing informed search method that accounts for the desirability and validity of designs during the search, called the A* algorithm~\cite{russell2003artificial}.

\subsection{A* search}
A* is a widely used algorithm for search-based problems. It works
in a best-first search manner by constructing and traversing the nodes in the tree that are most promising. Specifically, A* chooses nodes that minimize the cost function --
\begin{equation}
f(N) = g(N) + h(N)\,,
\label{eqn:total}
\end{equation}

where $g(N)$ represents the current cost of the design $\mathcal{D}_N$ at node $N$, and $h(N)$ corresponds to a heuristics that estimates the cost of parts that are required to be added to $\mathcal{D}_N$ for generating desired designs (corresponding to goal nodes), which can execute the user-specified motion. In other words, heuristics encapsulate the deviation of $\mathcal{D}_N$ from the goal design. %To drive A* towards as simple as possible yet valid designs, we define $g(N)$ to capture the design complexity, and define $h(N)$ to measure the design's validity for the task at hand.
Since a design with fewer number of components is economical and simpler to control and fabricate, we define $g(N)$ to be proportional to the number of parts in $\mathcal{D}_N$.

\begin{align}
g(N) &= \delta(\mathcal{D}_N)\,, \nonumber \\
h(N) &= \mathcal{E}_{IK}(\mathcal{D}_N, \mathcal{T})\,,
\label{eqn:searchcost}
\end{align}

where $\delta(\mathcal{D}_N)$ computes the number of parts such as actuators and links in the design $\mathcal{D}_N$ at node $N$. One can customize $\delta(\mathcal{D}_N)$ to penalize parts individually by using weights corresponding to a metric of choice. For instance, by using weights corresponding to the cost of components, one can drive the search to generate designs with fewer number of costly components such as actuators. To compute the heuristics $h(N)$, we evaluate the design's ability in executing a user-specified motion trajectory $\mathcal{T} \in SE(3)$ using Inverse-Kinematics (IK). We use IK to compute the required poses of the robot design so that the robot's end-effector can follow $\mathcal{T}$ as closely as possible. A higher IK error $\mathcal{E}_{IK}$ relates to the need of adding more parts to the current design $\mathcal{D}_N$, and thereby shows the deviation of $\mathcal{D}_N$ from desired designs.

For computing $\mathcal{E}_{IK}$, we first discretize $\mathcal{T}$ in time to obtain a set of $n$ target frames. Let $\textbf{q}$ represent an arbitrary pose of the robot. A robot design $\mathcal{D}$ is able to achieve the user-specified motion if there exists a set of poses $\textbf{q}_i (1 \leq i \leq n)$ of the robot arm that enable the arm's end-effector $e$ to achieve the target pose defined by $\mathcal{T}_i$ at $i^{th}$ frame without collisions. This Inverse-Kinematics (IK) computation is thus solved as an optimization over robot arm's poses $\textbf{q}$. 

\begin{equation}
\mathcal{E}_{IK}(\mathcal{D}, \mathcal{T}) = \underset{\textbf{q}_i}\min \sum_{i}^{n} \Delta(\mathcal{T}_i,e_i)^2\,
\label{eqn:ik}
\end{equation}
where $\Delta(\mathcal{T}_i,e_i)$ represent the error between the pose of robot arm's end-effector $e_i$ and the target pose specified by $\mathcal{T}_i$ at $i^{th}$ frame. $\textbf{q}_i$ represents the pose of the robot design $\mathcal{D}$ at $i^{th}$ frame. $\mathcal{E}_{IK}$ is the net IK error of the design $\mathcal{D}$ while following the trajectory $\mathcal{T}$. A video showing the A* performing such a search for robot arm structures is available \href{https://youtu.be/MOrijZu47EA}{here}\footnote{A* in action: \url{https://youtu.be/MOrijZu47EA}}.

\subsection{Admissibility of the heuristics}
Our IK error-based heuristic function (eq.~\ref{eqn:searchcost},~\ref{eqn:ik}) requires careful execution with regards to two issues. First, the IK error requires each design $\mathcal{D}$ to have an end-effector attached at the rear end of the robot. End-effectors are special parts in the library such as welding machine, sealing gun etc. that are user-specified and serve a specific purpose in the task. They may not be compatible with all the other parts in the library. As a result, a design $\mathcal{D}_N$ at an intermediate node in the tree may be composed of parts that do not allow end-effector connection. Secondly, 
for the efficiency of A* search process, the heuristics $h(N)$ needs to be \textit{admissible}. Without an admissible heuristics, which underestimates the actual cost of a node $N$, the search might traverse more nodes resulting in sub-optimal performance. 

To deal with these issues, we define a \textit{virtual} end-effector part in the library that is compatible and can attach with all other parts in the library. This not only allows computation of IK error for any intermediate designs in the tree, but also ensures admissibility. Any intermediate designs with a virtual end-effector is not feasible in the real world, and thereby its cost is an underestimate of the actual design's cost. The search termination criteria ensures that the goal nodes, which correspond to the desired designs, consist of the user-specified end-effector instead of the virtual one. 

\section{Preliminary results}
\label{sec:results}
\begin{figure}[htbp]
	\centering
	\includegraphics[width=\columnwidth]{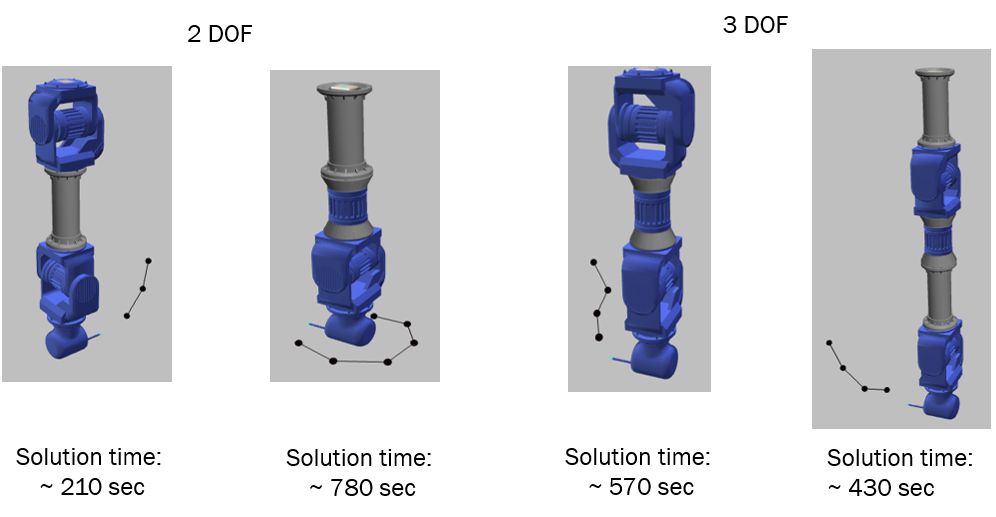}
	\caption{Various robotic arms with different DOF synthesized automatically with our system are shown here, along with the time taken to generate them. Corresponding user-specified target motion trajectories are shown in black. Each design is composed of modular parts such as actuators, links and end-effector. The actuators and end-effector are highlighted in blue.}
	\label{fig:results1}
\end{figure}

We demonstrate the capability of our automatic design approach by synthesizing robotic arms with different degrees of freedom (DOF) corresponding to various trajectory following scenarios. Figure~\ref{fig:results1} shows some of these designs generated by our system, given a robot base position and a motion trajectory to follow in an unconstrained environment. The time taken to generate these designs depend upon the complexity of target motion trajectories (length, required DOF). Apart from complexity of the target trajectory, environmental constraints also increase the design time. 

\begin{figure}[htbp]
	\centering
	\includegraphics[width=\columnwidth]{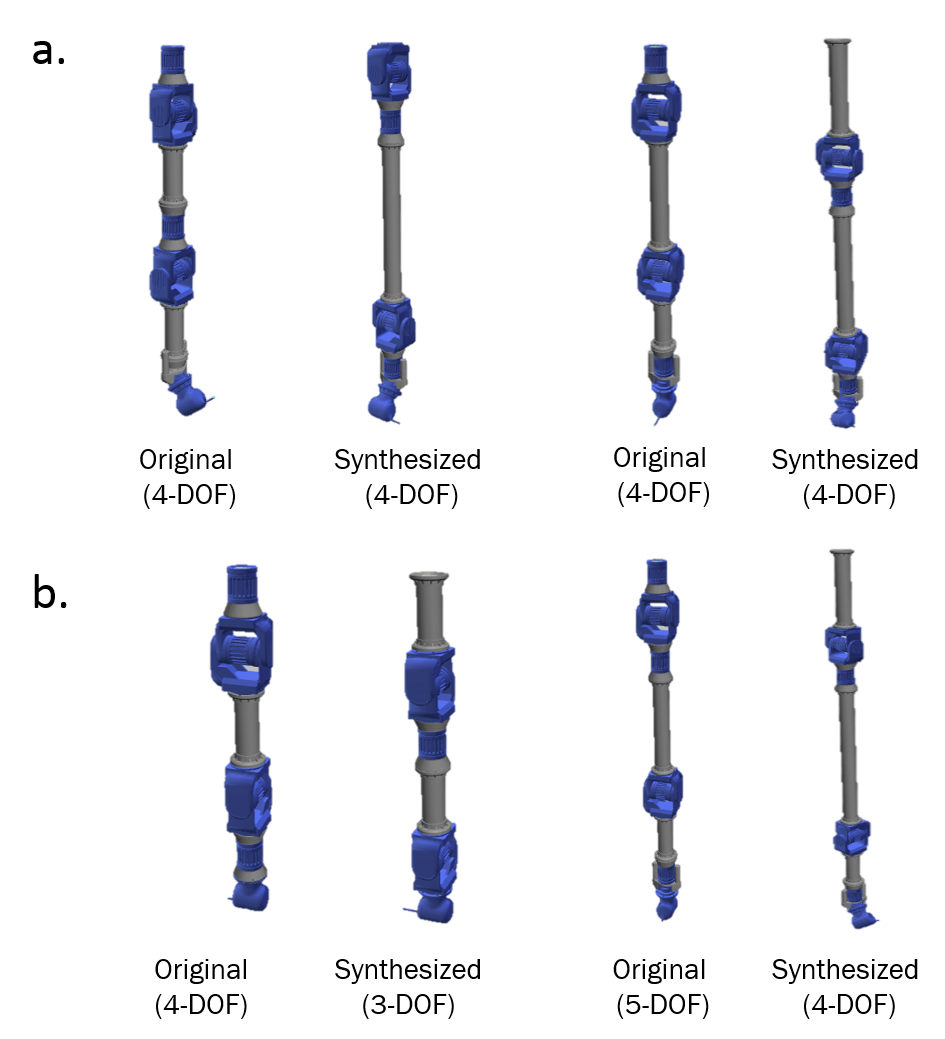}
	\caption{(a)~To validate our approach, we synthesize arm designs to follow trajectories that correspond to manually created designs (denoted as original designs). Our approach not only generates valid designs in all cases, but also finds simpler designs (with lower DOF) to follow trajectories that were originally generated using robotic arms with higher DOF in some cases, as shown in~(b). Actuators and end-effector are highlighted in blue.}
	\label{fig:results2}
\end{figure}

In order to validate the designs generated by our system, we manually created various robotic arm designs and corresponding motion trajectories. We then used these trajectories as targets for our automatic design generation. Comparison of the automatically generated designs with the original designs further highlight the strengths of our approach. We find that our system is able to find simpler designs (with fewer DOF) as compared to the original in some cases, especially when the environment is unconstrained (see fig.~\ref{fig:results2}(b)).

\section{Conclusion and Future work}
We proposed an automatic robot design approach for creating task-specific custom robots using reconfigurable and modular building blocks, given high-level descriptions of the target task. We demonstrated the capabilities of our approach by synthesizing robotic arms that execute user-specified motions. We also presented an interactive design environment that allowed users to define their task requirements and to intuitively interact with the automatic design synthesis. Our preliminary results show the ability of our system in synthesizing valid, and as simple as possible arm designs. Currently, the target motion trajectories for robot are user-specified. To enable design in much complex scenarios with large number of obstacles, we plan to integrate a motion planner in our system. We will also test our system to synthesize designs for multiple target trajectories in the future. Finally, we will apply our automatic design approach for synthesizing other robotic systems, to test its generality.

\bibliographystyle{IEEEtran}
\bibliography{references}

\end{document}